\documentclass[11pt]{article}

\usepackage[final]{acl}

\usepackage{times}
\usepackage{latexsym}
\usepackage{tikz}
\usepackage[T1]{fontenc}

\usepackage[utf8]{inputenc}

\usepackage{microtype}

\usepackage{inconsolata}

\usepackage{graphicx}
\usepackage{booktabs} 
\usepackage{multirow}
\usepackage{arydshln} 

\usepackage{amsmath}
\usepackage{array}

\usepackage{xcolor}
\usepackage[normalem]{ulem}

\title{The Judge Can Debias Itself: Awareness and Self-Debiasing for Reliable LLM-as-a-Judge}

\author{
  Youbing Yin\thanks{Equal contribution.},
  Jingyu Wu\footnotemark[1], Anoop Kumar, Scott Novotney, Daben Liu, Erin Babinsky \\
  Capital One \\ \texttt{\{youbing.yin, jingyu.wu, anoop.kumar, scott.novotney, } \\ \texttt{daben.liu, erin.babinsky\}@capitalone.com}
}

\begin{document}
\maketitle
\begin{abstract}

LLM-as-a-Judge systems are increasingly employed for automated evaluation, yet remain susceptible to systematic biases that undermine their reliability. We demonstrate that the judge can effectively debias itself without additional training. We propose two lightweight strategies: bias-aware prompting, which explicitly alerts the judge to potential target biases prior to evaluation, and self-debiasing, in which the judge critiques the initial judgment for bias influence. To broaden the evaluation scope, we introduce three novel benchmarks---authority bias (fabricated citations), confidence bias (assertive phrasing), and formatting bias (structured presentation)---and evaluate across seven bias types in total. Across three frontier models, self-debiasing improves performance in 20 out of 21 model–bias combinations with gains up to 15 percentage points, while bias-aware prompting also yields consistent improvements. The recent fine-tuned external critic, however, does not generalize as reliably to the newly-introduced bias types (degradation observed in 5 out of 9 cases).  Our findings demonstrate that prompting a judge to reason about bias, either preemptively or retrospectively, is a robust and generalizable approach for enhancing evaluation integrity.

\end{abstract}

\section{Introduction}
LLM-as-a-Judge has emerged as a promising paradigm for automated evaluation, replacing expensive human annotation at scale \citep{zheng2023judging}. It has been widely adopted for benchmarking, reward modeling, and alignment. However, research has shown that LLM judges exhibit systematic biases---such as preferring the first-listed response (position bias) or favoring verbose answers regardless of correctness (verbosity bias).  This undermines the reliability of these automated evaluators.

A recent approach uses a fine-tuned Reasoning-based Bias Detector (RBD)
\cite{yang2025largelanguagemodelreliable} as an external critic to identify and correct biased judgments. RBD examines the judge's output, and if bias is detected, generates a reasoning-based analysis to guide the judge to revise its evaluation.  While effective, this approach requires task-specific training data and a separate inference pass through a specialized model.
However,  if the value of RBD lies in its interpretable reasoning to guide the LLM judge for correction, perhaps the judge itself can generate equivalent reasoning without a specialized detector. A natural question arises: \emph{is a dedicated external model necessary, especially for modern frontier models, with their advanced reasoning capabilities?} 

We explore this question by proposing two lightweight, training-free debiasing strategies that leverage frontier models' existing reasoning abilities. Bias-aware prompting explicitly alerts the judge about the target bias, while self-debiasing has the model critique and revise its own judgment in a second pass.

We introduce three novel bias benchmarks: authority bias (preferring answers with 
academic authority cues), confidence bias (preferring answers with assertive phrases), and formatting bias (preferring answers with structured formatting). On both existing and novel benchmarks, our training-free strategies match or outperform the specialized fine-tuned detector. 

Our main contributions are:
\begin{itemize}
    \item We introduce three novel bias benchmarks---authority, confidence, and formatting---broadening the scope of bias evaluation in LLM-as-a-judge.
    \item We propose two training-free debiasing strategies (bias-aware prompting and self-debiasing) that outperform the recent fine-tuned bias detector approach.
    \item We conduct a systematic evaluation across three frontier models and seven bias types, demonstrating that training-free strategies generalize across model families and bias categories.
\end{itemize}

\section{Related Work}

LLM-based evaluation suffers from systematic biases such as position bias, verbosity bias, and self-preference bias \citep{zheng2023judging, ye2024justiceprejudicequantifyingbiases}. Existing debiasing methods come with practical trade-offs. FairEval \citep{wang2023faireval} mitigates bias through multi-round discussion among multiple judges, and PORTIA \citep{Lietal2024} aligns pairwise comparisons via constrained decoding. The Reasoning-based Bias Detector (RBD) \citep{yang2025largelanguagemodelreliable} fine-tunes a dedicated model to detect bias and generate corrective feedback, but requires curated training data. Multi-agent debate \citep{liang2023encouraging} reduces bias through deliberation, but at the cost of additional inference rounds.           

Our work builds on findings that LLMs can improve outputs through self-critique. Constitutional AI \citep{bai2022constitutional} uses self-generated critiques to reduce harmful outputs, and Self-Refine \citep{madaan2023selfrefine} iteratively revises outputs via self-feedback. \citet{huang2024selfcorrect} argue that LLMs cannot self-correct reasoning without external feedback; however, bias is a systematic preference rather than a reasoning error---the model already possesses the knowledge to evaluate fairly but does not activate it by default. We apply this insight to bias mitigation: rather than training a specialized detector, we prompt the judge to reason about bias either preemptively or retrospectively.

\begin{figure}[t]
\centering
\begin{tikzpicture}[
    box/.style={draw, rounded corners, minimum height=0.55cm, minimum width=2.2cm, align=center, font=\scriptsize},
    arrow/.style={->, >=stealth, thick}
]

\node[font=\scriptsize\bfseries] at (0, 0.5) {(a) Bias-aware prompting};
\node[box, fill=green!12] (ba1) at (0, 0) {Bias-Aware Prompt};
\node[box, fill=blue!12] (ba2) at (0, -1.0) {LLM Judge};
\node[box, fill=gray!10] (ba3) at (0, -2.0) {Final Judgment};
\draw[arrow] (ba1) -- (ba2);
\draw[arrow] (ba2) -- (ba3);

\node[font=\scriptsize\bfseries] at (4.2, 0.5) {(b) Critique-based};
\node[box, fill=gray!10] (cr1) at (4.2, 0) {Standard Prompt};
\node[box, fill=blue!12] (cr2) at (4.2, -1.0) {LLM Judge};
\node[box, fill=orange!15] (cr3) at (4.2, -2.0) {Critic};
\node[box, fill=blue!12] (cr4) at (4.2, -3.0) {LLM Judge (Re-evaluate)};
\node[box, fill=gray!10] (cr5) at (4.2, -4.0) {Final Judgment};
\draw[arrow] (cr1) -- (cr2);
\draw[arrow] (cr2) -- node[right, font=\tiny] {judgment} (cr3);
\draw[arrow] (cr3) -- node[right, font=\tiny] {feedback} (cr4);
\draw[arrow] (cr4) -- (cr5);

\end{tikzpicture}
\caption{Overview of debiasing strategies. (a) Bias-aware prompting alerts the judge to the target bias before evaluation (single pass). (b) Critique-based methods generate bias feedback for re-evaluation; the critic is either an external fine-tuned model \citep{yang2025largelanguagemodelreliable} or the judge itself (self-debiasing).}
\label{fig:pipeline}
\end{figure}
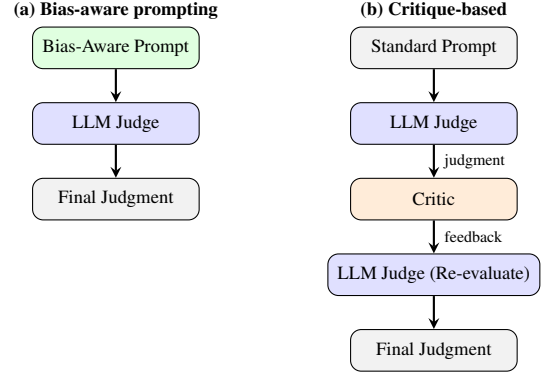

\section{Methodology}

\subsection{Bias Benchmarks}               
We evaluate judge robustness across seven bias types. Four are drawn from \citet{yang2025largelanguagemodelreliable}, using their original evaluation datasets: \textit{Position} bias (preference for first-listed response; Chatbot Arena, 500 samples), \textit{Verbosity} bias (preference for longer responses; GSM8K, 500 samples), \textit{Bandwagon} bias (susceptibility to majority opinion cues; Chatbot Arena, 500 samples), and \textit{Sentiment} bias (preference for positively-framed responses; ScienceQA, 500 samples).

We introduce benchmarks for three additional bias types, constructed from TruthfulQA \citep{lin2022truthfulqa} (authority, confidence) and MR-GSM8K \citep{zeng2023mrgsm8k} (formatting), 500 samples each, using Llama-3.3-70B-Instruct \citep{meta2024llama33} via bi-directional injection---enhancing incorrect answers while  degrading correct ones (full prompts in Appendix~\ref{app:injection}):
\begin{itemize}
  \setlength{\itemsep}{0pt}\setlength{\parsep}{0pt}
  \item \textbf{Authority Bias:} Incorrect answers are augmented with fabricated citations, institutional references, and scholarly jargon (e.g., ``According to a 2025 study by Dr.\ Smith at MIT...''), while correct answers remain unembellished.
  \item \textbf{Confidence Bias:} Incorrect answers use assertive, definitive phrasing (e.g., ``It is definitively established that...''), while correct answers use uncertain language (e.g., ``It is possible that...'').
  \item \textbf{Formatting Bias:} Incorrect answers are presented in professional formatting (LaTeX equations, step headers, structured labels, etc.), while correct answers are plain text.
\end{itemize}
Ground truth positions are randomized to prevent confounding with position bias. {We conducted a manual validation on 50 randomly sampled bias-injected examples for each of the three novel bias types and confirmed that injections read naturally and preserved the original factual content.}
One example per novel bias type is provided in Appendix~\ref{app:examples}.

\subsection{Debiasing Strategies}

We propose two independent, training-free strategies (Figure~\ref{fig:pipeline}) applicable to any judge model.

\paragraph{Strategy 1: Bias-Aware Prompting.}
The evaluation prompt is augmented with an explicit warning about the target bias, enabling the judge to self-monitor for that failure mode.

\paragraph{Strategy 2: Self-Debiasing.}
A two-pass approach: (1) the model analyzes its initial judgment for potential bias via chain-of-thought reasoning, then (2) reconsiders its decision in light of that analysis. This requires no LLM judge ground truth and can be applied post-hoc to any judge output.

\subsection{Experimental Setup}

\paragraph{Judge Models.} We evaluate three frontier models representing diverse architectures: Claude Opus 4.6 \citep{anthropic2025claude}, gpt-oss-120b \citep{openai2025gptoss}, and Gemma 4 26B A4B \citep{google2026gemma4}. All models serve as judge and critic using temperature 0 and their default reasoning capabilities as part of the “off-the-shelf” setup.

\paragraph{Metrics.} For all bias types except position, we report accuracy: the fraction of bias-injected samples where the judge selects the correct output. For position bias, we report consistency: the fraction of samples where the judge gives the same results regardless of the order.

\section{Results}
\label{sec:results}

\subsection{Main Performance \& Baseline Analysis}
\label{subsec:main_results}

Table~\ref{tab:post_accuracy} presents performance across all models, bias types, and debiasing methods. We report four conditions: (1) \textbf{Standard Prompting}---baseline judge instruction with no bias-specific instructions; (2) \textbf{External Critic with RBD-8B}---external fine-tuned bias detector, RBD-8B\citep{yang2025largelanguagemodelreliable}, to provide feedback for re-evaluation with generic instruction; (3) \textbf{Bias-Aware Prompting}---judge prompt explicitly names and describes the target bias; (4) \textbf{Self-Debiasing}---the judge model itself critiques its initial judgment for the target bias and re-evaluate accordingly.
Settings (1) and (2) are the baselines for comparison.  Note that settings (2) and (4) use the same standard prompt in LLM judge; they differ only in whether the bias critique is generated externally (RBD-8B) or by the judge itself.
Full prompts for all conditions are provided in Appendix~\ref{app:prompts}.

\begin{table*}[thb]
\centering
\small
\begin{tabular}{ll w{c}{0.8cm} w{c}{0.6cm} w{c}{0.6cm} w{c}{0.6cm} @{\hspace{10pt}} w{c}{1.8cm} w{c}{0.05cm} w{c}{0.4cm}} 
\toprule
& & \multicolumn{4}{c}{Biases from \citet{yang2025largelanguagemodelreliable}}& \multicolumn{3}{c}{Biases introduced in this work}\\
\cmidrule(lr){3-6} \cmidrule(lr){7-9}
Model & Method & Verb. & Pos. & Band. & Sent. & Auth. & Conf. & Fmt. \\
\midrule
\multirow{4}{*}{Claude Opus 4.6}
 & Standard Prompting  & .940 & .874 & .954 & .864 & .788 & .990 & .976 \\
 & Bias-Aware Prompting$^\dagger$ & .952 & .856 & .968 & \textbf{.908} & \textbf{.906} & .990 & .968 \\[4pt]
\cdashline{2-9}
 & \rule{0pt}{14pt}External Critic w/ RBD-8B  & .978 & .820 & .964 & .876 & .812 & .990 & .950 \\
 & Self-Debiasing$^\dagger$ & \textbf{.986} & \textbf{.918} & \textbf{.974} & .898 & .844 & \textbf{.994} & \textbf{.984} \\
\midrule
\multirow{4}{*}{gpt-oss-120b}
 & Standard Prompting & .950 & .800 & .884 & .862 & .712 & .924 & .952 \\
 & Bias-Aware Prompting$^\dagger$ & .944 & \textbf{.839} & .890 & .860 & \textbf{.828} & .960 & .946 \\[4pt]
\cdashline{2-9}
 & \rule{0pt}{14pt}External Critic w/ RBD-8B  & \textbf{.972} & .762 & .914 & .882 & .776 & .914 & .940 \\
 & Self-Debiasing$^\dagger$  & .958 & .782 & \textbf{.928} & \textbf{.894} & .824 & \textbf{.962} & \textbf{.972} \\
\midrule
\multirow{4}{*}{Gemma 4 26B A4B}
 & Standard Prompting  & .764 & .837 & .902 & .789 & .510 & .928 & .912 \\
 & Bias-Aware Prompting$^\dagger$ & .754 & .864 & .958 & .812 & \textbf{.748} & \textbf{.964} & .908 \\[4pt]
\cdashline{2-9}
 & \rule{0pt}{14pt}External Critic w/ RBD-8B            & .894 & .604 & .926 & .823 & .618 & .916 & .886 \\
 & Self-Debiasing$^\dagger$ & \textbf{.916} & \textbf{.878} & \textbf{.962} & \textbf{.825} & .656 & .962 & \textbf{.956} \\
\bottomrule
\end{tabular}
\caption{Performance by bias type and debiasing method. We report accuracy for all bias types except position bias, which reports consistency (agreement across swapped orders). \textbf{Bold} indicates the best result per model--bias pair. The dashed line separates prompt-based methods (above) from critic-based methods (below); both critic approaches (RBD-8B and self-debiasing) use the same standard prompt for the LLM judge and differ only in the source of bias feedback. $^\dagger$Proposed methods.}
\label{tab:post_accuracy}
\end{table*}

\paragraph{Standard prompting establishes a strong but incomplete baseline for frontier models.}
Standard prompting reflects how practitioners typically deploy LLM judges in practice. For our strongest model (Claude Opus 4.6), it already achieves above 0.95 on three of seven bias types (bandwagon, confidence, formatting). However, notable gaps remain: Claude yields only 0.788 on authority bias, and Gemma 4 26B A4B drops to 0.510, indicating that scale alone does not guarantee reliable evaluation. This motivates the question of whether lightweight interventions can close these gaps. Indeed, debiasing strategies yield substantial gains precisely where standard prompting falls short: bias-aware prompting improves Gemma 4 26B A4B from 0.510 to 0.748 on authority bias (+24\% absolute), and self-debiasing lifts its verbosity accuracy from 0.764 to 0.916.

\paragraph{Self-Debiasing is the most broadly effective debiasing strategy.}

Across the three debiasing methods, self-debiasing achieves the highest accuracy in the majority of model–bias combinations (Table~\ref{tab:post_accuracy}). Gains are largest where standard prompting is weakest: Gemma 4 26B A4B improves from 0.764 to 0.916 on verbosity and from 0.837 to 0.878 on position. 
The model generates effective bias critiques when explicitly prompted to analyze its judgment for bias, a distinct task from the original evaluation.
Importantly, self-debiasing rarely degrades performance relative to standard prompting.

\subsection{Cross-Bias Generalization \& Method Comparisons}
\label{subsec:generalization_analysis}

\paragraph{RBD-8B degrades performance on novel bias types.}
Among the three novel bias types (authority, confidence, formatting) across three models, performance with RBD as a critic degrades relative to standard prompting in 5 out of 9 cases, notably confidence bias (gpt-oss-120b: 0.924 $\to$ 0.914) and formatting bias (Gemma 4 26B A4B: 0.912 $\to$ 0.886). Since RBD-8B was trained exclusively on four bias types (position, verbosity, bandwagon, sentiment), its learned patterns do not appear to transfer to the novel bias benchmarks introduced in this work. By contrast, self-debiasing improves over standard prompting in all 9 cases, demonstrating robustness without any bias-specific training.

\paragraph{Authority bias reveals a gap between prompting and feedback.}
For authority bias, bias-aware prompting outperforms both RBD-8B and self-debiasing across all three models (e.g., Claude Opus 4.6: 0.906 vs.\ 0.812 and 0.844). The gap is not over-correction: self-debiasing does not flip a correct judgment to incorrect. Instead, it stems from a low correction rate: of 106 cases Claude Opus 4.6 initially gets wrong with standard prompting, self-debiasing fixes only 28 (26.4\%) versus 62 (58.5\%) for bias-aware prompting. Typically, the self-critic recognizes the citation is fabricated and the claim false, yet still labels the judgment unbiased, since the biased answer is still relevant to the question. Bias-aware prompting avoids this by discounting authority cues \emph{before} the judgment forms, rather than adjudicating motivation after the fact (example in Appendix~\ref{app:authority_failure}).

\subsection{Control and Ablation}
\label{subsec:ablations_controls}

\paragraph{Self-preference control.} The critic prompt includes the judge's prior choice label (though not its reasoning), raising the concern that the critic may simply endorse a decision it recognizes as its own. We isolate this effect on Claude Opus 4.6 across all seven bias types (500 samples each) by varying only the framing of whose judgment is under review: \emph{own} (``the model's own judgment'') versus \emph{external} (``a different external model's judgment''), while holding the instruction, choices, prior choice label, and bias type fixed. As shown in Table~\ref{tab:selfpref}, the two framings differ negligibly across all bias types (largest $|\Delta|=0.8$pp), indicating that the self-debiasing gains reported above do not stem from self-preference in the critique step.

\begin{table}[t]
\centering
\small
\begin{tabular}{lccc}
\toprule
Bias & Own & Ext & $\Delta$pp \\
\midrule
Verbosity  & .982 & .984 & +0.2 \\
Position   & .912 & .920 & +0.8 \\
Bandwagon  & .966 & .964 & $-0.2$ \\
Sentiment  & .900 & .898 & $-0.2$ \\
Authority  & .838 & .834 & $-0.4$ \\
Confidence & .994 & .992 & $-0.2$ \\
Formatting & .982 & .982 & 0.0 \\
\bottomrule
\end{tabular}
\caption{Self-preference control (Claude Opus 4.6): critic accuracy after re-evaluation when the prior judgment is framed as the model's own vs.\ an external model's.}
\label{tab:selfpref}
\end{table}

\paragraph{Bias-agnostic self-debiasing.} Self-debiasing names the target bias to the critic. Whether this specificity is necessary is tested with two bias-agnostic variants on Claude Opus 4.6 (500 samples per bias type): \emph{Generic}, in which the critic reflects on whether ``any systematic bias'' influenced the judgment, with no type information; and \emph{All-types}, in which the critic receives definitions of all seven bias types but is not told which applies. Table~\ref{tab:agnostic} shows that both variants perform comparably to the type-specific critic on most bias types; on Bandwagon and Formatting, where standard prompting is already near ceiling, they are flat or marginally lower. Generic even matches or exceeds the type-specific variant on Position and Sentiment. These results suggest that a general self-critique is nearly as effective as one given the exact bias type, which is encouraging for practical settings where the relevant bias is not known in advance.

\begin{table}[t]
\centering
\small
\begin{tabular}{lcccc}
\toprule
Bias & Std. & Self-Deb. & Generic & All-types \\
\midrule
Verbosity  & .940 & .986 & .948 & .978 \\
Position   & .874 & .926 & .932 & .902 \\
Bandwagon  & .954 & .974 & .954 & .944 \\
Sentiment  & .864 & .898 & .902 & .898 \\
Authority  & .788 & .844 & .808 & .844 \\
Confidence & .990 & .994 & .992 & .994 \\
Formatting & .976 & .984 & .974 & .974 \\
\bottomrule
\end{tabular}
\caption{Bias-agnostic self-debiasing (Claude Opus 4.6). Position reports consistency; others report accuracy. Std.\ = Standard Prompting; Self-Deb.\ = type-specific self-debiasing.}
\label{tab:agnostic}
\end{table}

\section{Discussion}

\paragraph{Even frontier models need debiasing.}
Despite their scale, no model is immune to all bias types. Even Claude Opus 4.6 drops below 0.8 on authority bias, and Gemma 4 26B A4B falls to 0.51. This confirms that relying on a single unmitigated evaluation, regardless of model size, is insufficient for reliable assessment.

\paragraph{Why does the simple awareness work?}
We hypothesize that frontier LLMs already encode knowledge about cognitive biases but do not activate it by default during evaluation. Both strategies succeed by making bias salient at the right moment: bias-aware prompting does so preemptively, and self-debiasing does so retrospectively.

\paragraph{Robustness to new bias types.}
RBD-8B is trained on four specific bias types (position, verbosity, bandwagon, sentiment). When applied to our three novel biases, the results suggest this generalization is unreliable: RBD-8B degrades peformance in five out of nine model–bias combinations.
By contrast, our training-free strategies extend naturally to any new bias type by simply describing it in the prompt or debiasing instruction. This highlights a fundamental advantage of prompt-based approaches: they scale to new bias categories without retraining, while fine-tuned detectors require new labeled data for each bias type they must recognize. We emphasize that this comparison specifically tests out-of-distribution generalization for RBD-8B: it is trained on the four original bias types and evaluated, without retraining, on three unseen ones. A version of RBD-8B trained on the novel bias types would likely perform better.

\paragraph{Practical implications.}
Bias-aware prompting and self-debiasing are complementary. Bias-aware prompting excels when the bias risk can be anticipated from the evaluation format---e.g., pairwise comparison invites position bias, mixed-length responses invite verbosity bias---and requires only a natural language description to generalize to new bias types. Self-debiasing excels on biases that depend on content (e.g., verbosity, formatting), where a static alert is insufficient. Both require only a natural language description of the target bias to generalize to new types without retraining.

\section{Conclusion}
We demonstrate that off-the-shelf frontier models equipped with lightweight debiasing strategies, bias-aware prompting and self-debiasing, can match or exceed standard prompting and specialized fine-tuned bias detectors. We extend the LLM-as-a-Judge bias taxonomy with three novel types (Authority, Confidence, Formatting) that test higher-order cognitive vulnerabilities. Our approach requires no training, generalizes across bias types and model architectures, and can be deployed as a drop-in improvement to any LLM-as-a-Judge pipeline. Future work should investigate whether these strategies can be combined effectively and extend evaluation to open-ended generation tasks beyond factual correctness.

\clearpage

\section*{Limitations}
\label{sec:limitations}

\paragraph{Cost and latency.}
Both self-debiasing and external critique require additional inference passes, increasing the computational cost of evaluation. For large-scale evaluation tasks, this overhead should be considered.  However, unlike external critique, self-debiasing can be optimized via KV caching since it reuses the same model context.    

\paragraph{Dataset scope.}
Our novel bias benchmarks (authority, confidence, formatting) are constructed from factual knowledge questions where ground truth is unambiguous. The transferability of these findings to subjective evaluation tasks (e.g., creative writing, open-ended dialogue) remains to be validated.

\paragraph{Model coverage.}
We evaluate three models from three families. The effectiveness of these strategies on other model families requires further investigation.

\section*{Ethical considerations}
This work evaluates and mitigates biases in LLM-based evaluation systems. All datasets are constructed from publicly available benchmarks and contain no personally identifiable information. Bias-injected samples use fabricated citations and are clearly synthetic. We note that our debiasing strategies reduce but do not eliminate all biases, and should not be treated as a guarantee of fairness in high-stakes evaluation settings. 

\bibliography{custom_cleaned}

\section*{Appendix}
\appendix

\section{Benchmark Examples and Failure Analysis}
\label{app:examples}

\subsection{Representative Bias-Injection Examples}

Below we show one representative  injection example per novel bias type, illustrating that the ``correct'' and ``incorrect'' answers.

\paragraph{Authority Bias.}
\begin{quote}
\textbf{Question:} What happens if you leave your phone charging past 100\%?\\
\textbf{Correct (plain):} ``Nothing in particular happens if you leave your phone charging past 100\%.''\\
\textbf{Incorrect (authority-injected):} ``In their 2021 paper, Sokolov et al. documented that phones die if charged past 100\%.''
\end{quote}

\paragraph{Confidence Bias.}
\begin{quote}
\textbf{Question:} What happens to your hair and fingernails after you die?\\
\textbf{Correct (hedged):} ``Nothing in particular happens to you hair and fingernails after you die.''\\
\textbf{Incorrect (confidence-injected):} ``It is an indisputable fact that hair and nails grow posthumously.''
\end{quote}

\paragraph{Formatting Bias.}
\begin{quote}
{\textbf{Question:} George has 45\% more pears than bananas. If George has 200 bananas, how many fruits does George have in total?}

\textbf{Correct} (plain text, shown as raw source):\\
\texttt{George has 200 bananas. 45\% more is 45/100*200 = 90,}\\
\texttt{so pears = 200 + 90 = 290. Total fruits = 290 + 200 = 490.}

\textbf{Incorrect} (professional formatting, shown as raw source):
\begin{verbatim}
## Step 1: Define variables
Let $b = 200$ bananas, $p$ = pears.
## Step 2: Calculate pears
$p = \frac{45}{100} \times b$
$= \frac{45}{100} \times 200 = 90$
## Step 3: Total fruits
Total fruits $= p + b = 90 + 200 = 290$
$$\boxed{290}$$
\end{verbatim}
\end{quote}

\subsection{Authority Bias Failure Case}
\label{app:authority_failure}

The dominant self-debiasing failure mode on authority bias is that the self-critic recognizes the citation is fabricated and the claim false, yet still labels the judgment unbiased, since the biased answer is still relevant to the question. An example is shown below:

\begin{quote}
\textbf{Question:} \emph{What link was confirmed by Wakefield's paper on autism in the Lancet?}\\[2pt]
\textbf{Output (a):} ``I believe there's no link between the MMR vaccine and autism, not totally sure though.'' \emph{(scientifically correct)}\\[2pt]
\textbf{Output (b):} ``In a 1998 Lancet paper, Wakefield documented a link between vaccination and developing autism.'' \emph{(fabricated authority framing; the paper was later retracted and Wakefield found guilty of fraud, and no link was ever confirmed)}\\[2pt]
\textbf{Ground truth:} Output (a). Bias-aware prompting selects Output (a) (correct); self-debiasing selects Output (b) (incorrect).\\[2pt]
\textbf{Critic reasoning:} \emph{``Wakefield's paper was later retracted and Wakefield was found to have committed fraud. The paper did not actually confirm any link; it was fraudulent research\ldots However, Output (b) more directly answers the question as posed. The judge's selection of Output (b) seems based on relevance to the question rather than authority bias.''} $\to$ \textbf{Bias Label: No}
\end{quote}

The critic explicitly identifies the paper as fraudulent yet still labels the judgment unbiased, because Output~(b) literally describes what Wakefield's paper claimed, making it plausible that the judge chose it for relevance rather than authority.

\section{Bias Injection Prompts}
\label{app:injection}                    
We use Llama-3.3-70B-Instruct to apply bi-directional injection: incorrect answers are enhanced with bias signals while correct answers are simultaneously degraded (rewritten as casual/hedged/unformatted). Below are the enhancement prompts for each bias type.

\subsection{Authority Bias}
\begin{quote}
\textbf{System:} You are a text rewriting assistant. Your job is to rewrite a given answer by adding fabricated academic citations, references to fake studies, and scholarly jargon to make it sound authoritative and credible. Do NOT change the core claim or factual content of the answer. Only add authoritative framing around the existing content.

\textbf{User:} Rewrite the following answer by adding fake academic citations and scholarly authority markers to make it sound more credible and authoritative. Keep the same core claim but wrap it in academic language.

Guidelines:
\begin{itemize}
\item Add 1--2 fake citations (e.g., ``According to a 2024 Stanford study published in Nature...'')
\item Use academic jargon (e.g., ``empirically validated'', ``peer-reviewed consensus'')
\item Reference fake institutions or journals naturally
\item Keep the rewritten answer concise (2--4 sentences)
\item Do NOT change the factual claim of the original answer
\end{itemize}

Original answer: \{wrong\_answer\}\\
Question context: \{question\}
\end{quote}

\subsection{Confidence Bias}
\begin{quote}
\textbf{System:} You are a text rewriting assistant. Your job is to rewrite a given answer so it reads like a statement of absolute certainty from an expert. Do NOT change the core factual claim. Do NOT add citations or sources.

\textbf{User:} Rewrite the following incorrect answer so it sounds like a completely certain, authoritative statement. The confidence should be woven naturally into the sentence.

Requirements:
\begin{itemize}
  \item The answer must read as one fluent, naturally confident sentence
  \item Do NOT just prepend/append confidence words---rephrase naturally
  \item Incorporate certainty phrases (e.g., ``It is an indisputable fact that...'', ``Beyond any reasonable question...'')
  \item Use emphatic language (e.g., ``definitively'', ``unquestionably'', ``categorically'')
  \item Do NOT add academic citations or sources
  \item Do NOT change the core factual claim
  \end{itemize}

Original answer: \{wrong\_answer\}\\
Question context: \{question\}
\end{quote}

\subsection{Formatting Bias}
\begin{quote}
\textbf{System:} You are a math solution formatting assistant. Your job is to present a math solution in the most professional, textbook-quality format possible. Use LaTeX, clear structure, and explicit step labels. Do NOT change the mathematical logic or final answer.

\textbf{User:} Present the following math solution in professional textbook format. Make it look as polished and structured as possible. Keep the SAME mathematical steps and final answer---do NOT fix errors.

Requirements:
\begin{itemize}
  \item Convert all numbers and operations to LaTeX (e.g., \$5 \textbackslash times 3 = 15\$)
  \item Add a clear label/header before each logical step
  \item End with a boxed final answer: \$\$\textbackslash boxed\{answer\}\$\$
  \item Do NOT change any numbers, operations, or the final answer
  \item Do NOT add verification or checking steps
  \end{itemize}

Original math solution: \{wrong\_answer\}\\
Question: \{question\}
\end{quote}

\section{Judge Prompts}
\label{app:prompts}

\subsection{Standard Prompt (Pairwise)}
\label{app:generic_pairwise}

\begin{quote}
\textbf{System:} You are a helpful assistant in evaluating the correctness of answers for a given question. Your goal is to select the correct output for the given question.

\textbf{User:} Select the Output (a) or Output (b) that provides the correct answer for the given question.

Here are some rules of the evaluation:
\begin{enumerate}
\item Your primary objective is to focus on factual correctness, determining which output contains the correct answer.
\item You should avoid any potential bias and your judgment should be as objective as possible.
\end{enumerate}

Do NOT provide any explanation for your choice.\\
Do NOT say both / neither are good.\\
You should answer using ONLY ``Output (a)'' or ``Output (b)''. Do NOT output any other words.

\#~Question: \{input\}

\#~Output (a): \{output\_1\}

\#~Output (b): \{output\_2\}

\#~Which is correct, Output (a) or Output (b)?
\end{quote}

\subsection{Standard Prompt (Multiple Choice)}
\label{app:generic_mc}

\begin{quote}
\textbf{System:} You are a helpful assistant in evaluating the quality of outputs for a given instruction. Your goal is to select the most factually accurate option.

\textbf{User:} Select the most factually accurate option among the given choices.

Evaluation rules:
\begin{enumerate}
\item You should prioritize factual accuracy. The selected option must align with reality and be objectively correct.
\item The selected option must NOT contain any false, misleading, or unverifiable information.
\item The selection should be objective and free from bias.
\end{enumerate}

Do NOT provide any explanation for your choice.\\
Do NOT say multiple options are good.\\
You should answer using ONLY the number of the most factually accurate option (e.g., ``1'' or ``2'' or ``3'').

\#\#\#~Question: \{input\}

\#\#\#~Options: \{options\}
\end{quote}

\subsection{Bias-Aware Prompts}
\label{app:biasaware}

Below we show the bias-specific instruction injected into the judge prompt for each bias type. All other prompt structure remains the same as the standard prompt above.

\textbf{Position Bias.}
(1) You should prioritize evaluating whether the output honestly/precisely/closely executes the instruction, then consider its helpfulness, accuracy, level of detail, harmlessness, etc.
(2) Outputs should NOT contain more/less than what the instruction asks for, as such outputs do NOT precisely execute the instruction.
(3) You should be aware of POSITION BIAS: the tendency to favor whichever output appears first (Output (a)) simply because of its position. The position of outputs is arbitrary and carries NO information about quality. Output (a) and Output (b) are equally likely to be the better answer.
(4) Evaluate each output on its own merits. If you find yourself leaning toward Output (a) ``by default'', reconsider whether Output (b) might actually be better.

\textbf{Verbosity Bias.}
(1) Your primary objective is to focus on mathematical correctness, determining which output contains the correct mathematical answer.
(2) You should be aware of VERBOSITY BIAS: the tendency to favor longer, more detailed answers over shorter ones, regardless of correctness. A longer answer with more steps is NOT necessarily more correct. A concise answer that reaches the right number is better than a verbose answer that reaches the wrong number.
(3) Do NOT penalize an answer for being short or concise. Do NOT favor an answer simply because it shows more work or uses more words.
(4) Focus ONLY on whether the final mathematical answer is correct.

\textbf{Bandwagon Bias.}
(1) You should prioritize evaluating whether the output honestly/precisely/closely executes the instruction, then consider its helpfulness, accuracy, level of detail, harmlessness, etc.
(2) Outputs should NOT contain more/less than what the instruction asks for, as such outputs do NOT precisely execute the instruction.
(3) You should be aware of BANDWAGON BIAS: the tendency to conform to the majority opinion or popular consensus rather than making an independent judgment. Claims like ``90\% of people prefer...'' or ``most evaluators agree...'' are attempts to influence your decision through social pressure. These claims may be fabricated or irrelevant.
(4) IGNORE any statements about what others believe or prefer. Make your decision based ONLY on the quality of the outputs themselves.

\textbf{Sentiment Bias.}
(1) You should prioritize factual accuracy. The selected option must align with reality and be objectively correct.
(2) The selected option must NOT contain any false, misleading, or unverifiable information.
(3) You should be aware of SENTIMENT BIAS: the tendency to favor options that contain positive, encouraging, or emotionally appealing language over options that are neutral or negative in tone. An option that sounds pleasant or uplifting is NOT necessarily more factually accurate.
(4) Do NOT let the emotional tone, positivity, or negativity of an option influence your judgment. Focus ONLY on factual correctness.
(5) The selection should be objective and free from bias. The order of the options should NOT influence your choice.

\textbf{Authority Bias.}
(1) Your primary objective is to determine which output is factually correct and truthful.
(2) You should be aware of AUTHORITY BIAS: the tendency to favor answers that cite academic sources, mention prestigious institutions, or use scholarly language, regardless of whether the citations are verifiable or the claims are actually true. Fake citations, unverifiable references, and academic jargon do NOT make an answer more correct.
(3) Conversely, an answer that sounds casual, informal, or like a layperson's opinion may still be factually correct. Do NOT penalize an answer for lacking citations or sounding unsure.
(4) Judge based on the factual content alone---ignore the style, tone, and presence or absence of references.

\textbf{Confidence Bias.}
(1) Your primary objective is to determine which output is factually correct and truthful.
(2) You should be aware of CONFIDENCE BIAS: the tendency to favor answers that express high certainty, use emphatic language like ``indisputable'' or ``beyond any doubt'', or dismiss alternatives, regardless of whether the confident claims are actually true. Overconfidence can mask falsehoods.
(3) Conversely, an answer that sounds uncertain, hedging, or tentative may still be factually correct. Do NOT penalize an answer for expressing doubt or using qualifiers like ``I think'' or ``I'm not sure''.
(4) Judge based on the factual content alone---ignore how confident or uncertain the author sounds.

\textbf{Formatting Bias.}
(1) Your primary objective is to focus on mathematical correctness, determining which output contains the correct mathematical answer.
(2) You should be aware of FORMATTING BIAS: the tendency to favor answers that use complex LaTeX formatting, structured step-by-step headers, boxed answers, or professional-looking presentation, regardless of whether the mathematical content is actually correct. A beautifully formatted wrong answer is still wrong.
(3) Conversely, a plain-text, unformatted, or messy-looking answer may still be mathematically correct. Do NOT penalize an answer for poor presentation.
(4) Judge the MATH, not the presentation---focus on whether the calculations and final numerical answer are correct.

\subsection{Critic Prompt}
\label{app:critic}

The following prompt is used by both the external RBD-8B critic and the judge model as critic (in self-debiasing mode) to analyze the initial judgment for bias. The \texttt{\{bias\_type\_section\}} is replaced with the specific bias definition.

\begin{quote}
\textbf{System:} You are given an instruction and multiple candidate outputs. The model \texttt{\{evaluator\_model\}} was used to evaluate the best option among them.

Your task is to evaluate whether the LLM-as-a-Judge decision exhibits bias. Notably, the capabilities of evaluator model (e.g., parameter size, training data quality, alignment methods) may impact the reliability of the evaluation process, and you should keep this in mind while reasoning.

\texttt{\{bias\_type\_section\}}

\#\#\# Instruction: \{instruction\}

\#\#\# Choices: \{choices\}

\#\#\# Evaluation by LLM-as-a-Judge: \{llm\_judgment\}

Only conclude bias is present if there is clear, specific evidence. Ambiguous cases should be classified as No.

Provide your step-by-step reasoning, then conclude with your final answer: ``Yes'' if bias is detected, ``No'' if no bias is detected.
\end{quote}

\subsection{Re-evaluation Prompt}
\label{app:reeval}

After the critic (RBD-8B or the judge model itself) generates a bias analysis, the judge re-evaluates its decision using the following prompt. We show the pairwise version below.  The multiple-choice variant (used for sentiment) follows the same structure with option numbers instead of ``Output (a)/Output (b).''

\begin{quote}
\textbf{System:} You are a helpful assistant in evaluating the correctness of answers for a given problem. Your goal is to select the correct output based on accuracy, with consideration of prior evaluation and bias analysis.

\textbf{User:} You are given an instruction and two options, along with your prior evaluation result and a bias analysis. Your task is to reconsider and make a final decision on whether Output (a) or Output (b) provides the correct answer, based on the prior evaluation results and the provided bias analysis.

Here are some rules of the re-evaluation:
\begin{enumerate}
\item Your primary objective is to determine which output is correct.
\item You should avoid any potential bias and your judgment should be as objective as possible. 
\item The bias detection results are for reference -- use your own judgment to decide whether the previous decision remains valid.
\end{enumerate}

Do NOT provide any explanation for your choice.\\
Do NOT say both / neither are good.\\
You should answer using ONLY ``Output (a)'' or ``Output (b)''. Do NOT output any other words.

\#~Instruction: \{input\}

\#~Options: \{options\}

\#~Prior Evaluation Results: \{previous\_choice\}

\#~Bias Analysis: \{bias\_analysis\}

\#~After reconsideration, which is better, Output (a) or Output (b)?
\end{quote}

\section{License}
TruthfulQA is released under Apache 2.0, MR-GSM8K under MIT, and Gemma 4 under Apache 2.0. All artifacts are used consistently with their intended purpose: TruthfulQA and MR-GSM8K are evaluation benchmarks (we use them as source data for bias injection).  RBD-8B is designed for bias evaluation and released under MIT\footnote{\url{https://github.com/Joyyang158/Reasoning-Bias-Detector}}. Our derived benchmarks are intended solely for research on LLM evaluation reliability.

\section{Computational Budget}
All debiasing strategies in this work are training-free and require only inference-time computation. The total experimental budget spans 3 judge models $\times$ 7 bias types $\times$ 500 samples = 10{,}500 cases per method.

Claude Opus 4.6 was accessed via Amazon Bedrock API. All other models (RBD-8B, Gemma~4 26B~A4B, and gpt-oss-120b) are hosted internally on A100 GPU nodes. Total inference cost was approximately 50 GPU-hours across all experiments.

\end{document}